\documentclass[final]{cvpr}

\usepackage{times}
\usepackage{epsfig}
\usepackage{graphicx}
\usepackage{amsmath}
\usepackage{amssymb}
\usepackage{stmaryrd}
\usepackage{makecell}
\usepackage[OT1]{fontenc} 
\usepackage{caption}
\usepackage{adjustbox}

\usepackage[pagebackref=true,breaklinks=true,colorlinks,bookmarks=false]{hyperref}

\DeclareMathOperator*{\argmin}{arg\,min}

\def\cvprPaperID{10645} 
\def\confYear{CVPR 2021}
\begin{document}

\title{MongeNet: Efficient Sampler for Geometric Deep Learning}
\author{L\'eo Lebrat$^{1,2\ \dag}$, Rodrigo Santa Cruz$^{1,2\ \dag}$, Clinton Fookes$^{2}$ and Olivier Salvado$^{1,2}$\\
$^{1}$ CSIRO, $^{2}$ Queensland University of Technology, $^{\dag}${equal contribution} \\
{\url{https://lebrat.github.io/MongeNet}}
}
\twocolumn[{%
\renewcommand\twocolumn[1][]{#1}%
\maketitle
\thispagestyle{empty}
\vspace{-1cm}\begin{center}
    \centering
    \includegraphics[width=.49\textwidth]{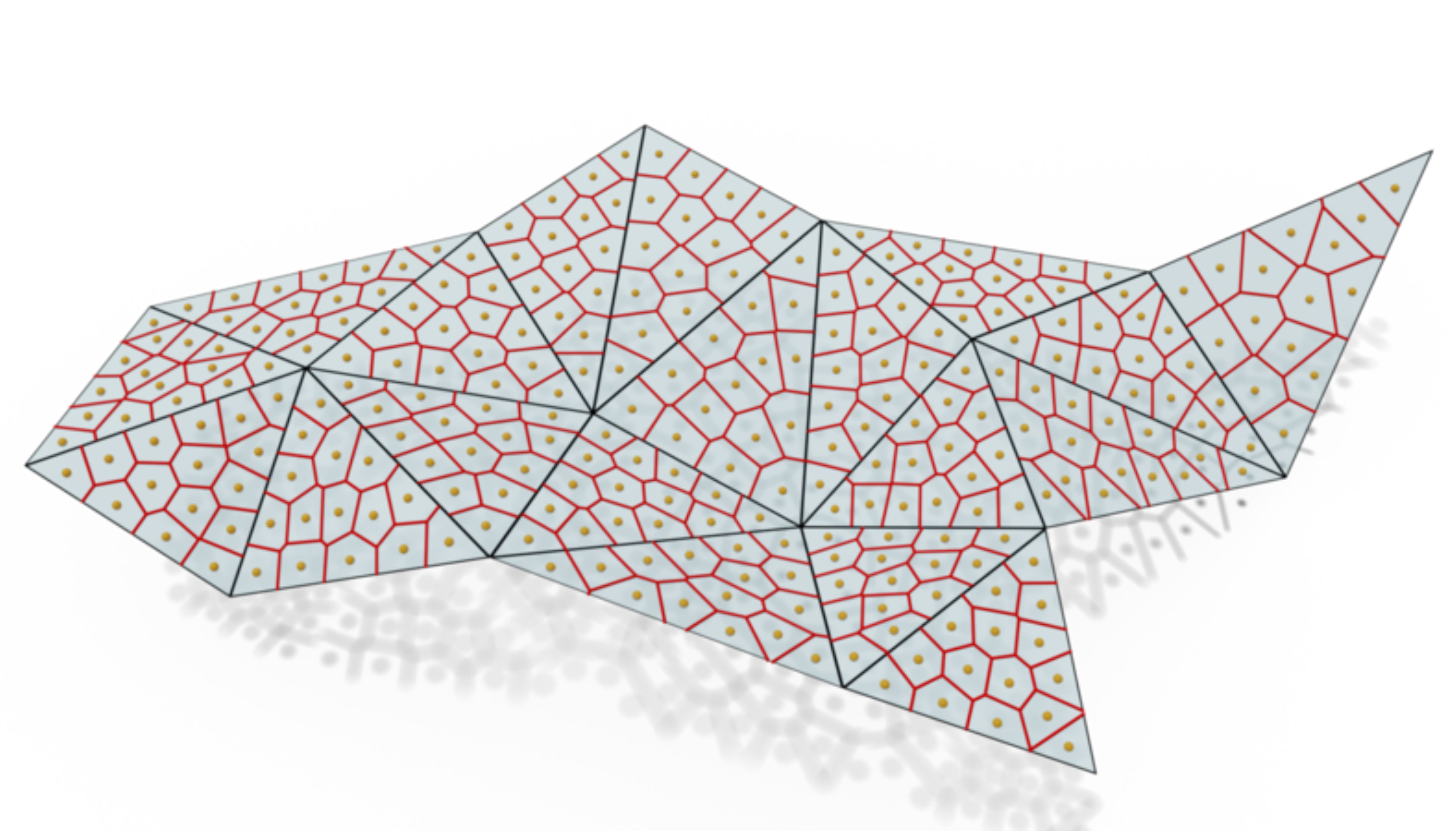}
    \includegraphics[width=.49\textwidth]{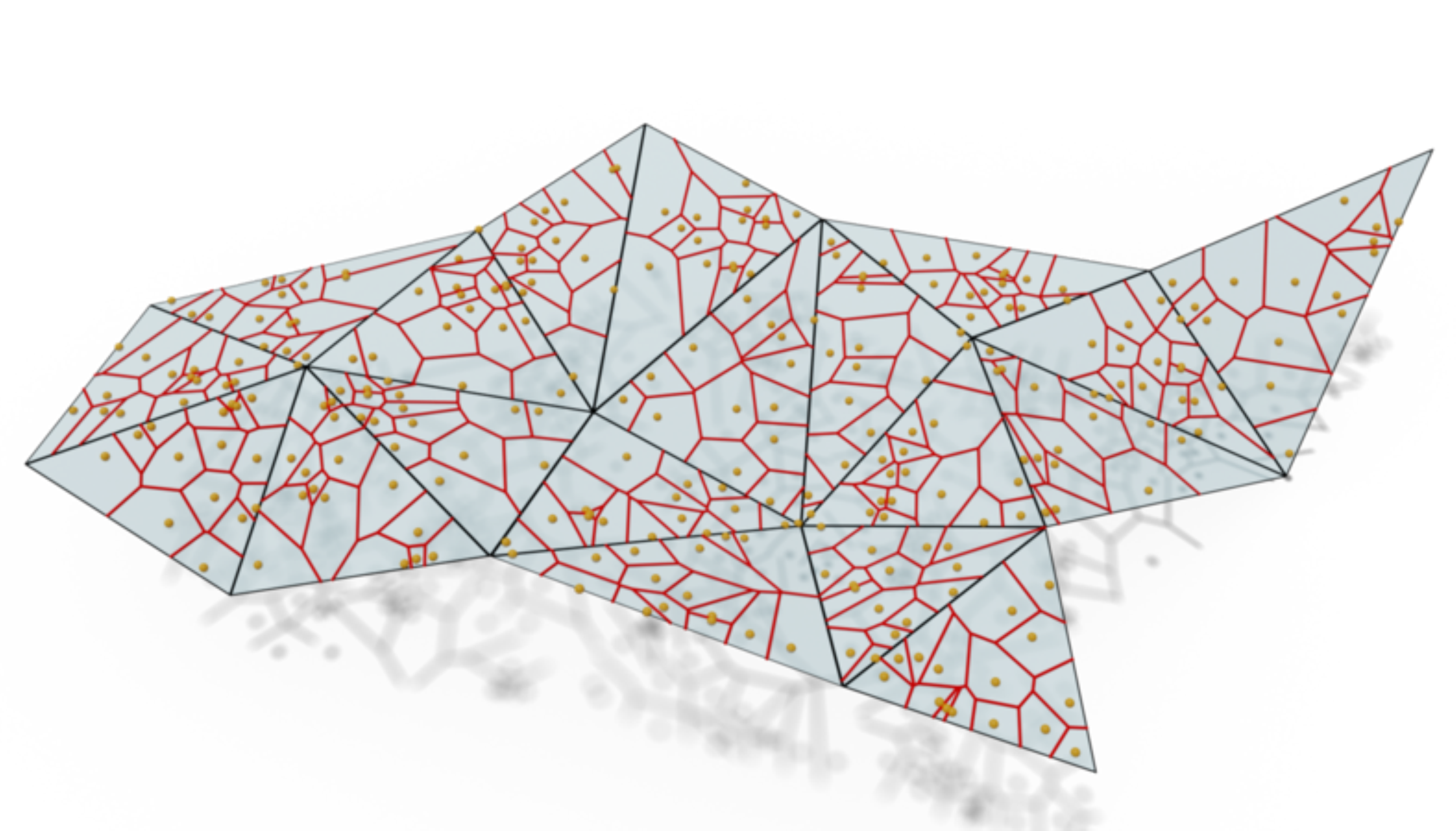}
    \captionsetup{type=figure}
    \captionof{figure}{Discretization of a mesh by a point cloud. \textbf{Left:} MongeNet discretization. \textbf{Right:} Classical random uniform discretization. The resulting Voronoi Tessellations spawn on the triangles of the mesh are shown in red.}\label{fig:mainPicturePaper}
\end{center}%
}]

\begin{abstract}
   Recent advances in geometric deep-learning introduce complex computational challenges for evaluating the distance between meshes. From a mesh model, point clouds are necessary along with a robust distance metric to assess surface quality or as part of the loss function for training models. Current methods often rely on a uniform random mesh discretization, which yields irregular sampling and noisy distance estimation. In this paper we introduce MongeNet, a fast and optimal transport based sampler that allows for an accurate discretization of a mesh with better approximation properties. We compare our method to the ubiquitous random uniform sampling and show that the approximation error is almost half with a very small computational overhead.
\end{abstract}


\section{Introduction}

\begin{figure*}[!ht]
    \centering
    \includegraphics[width=\textwidth]{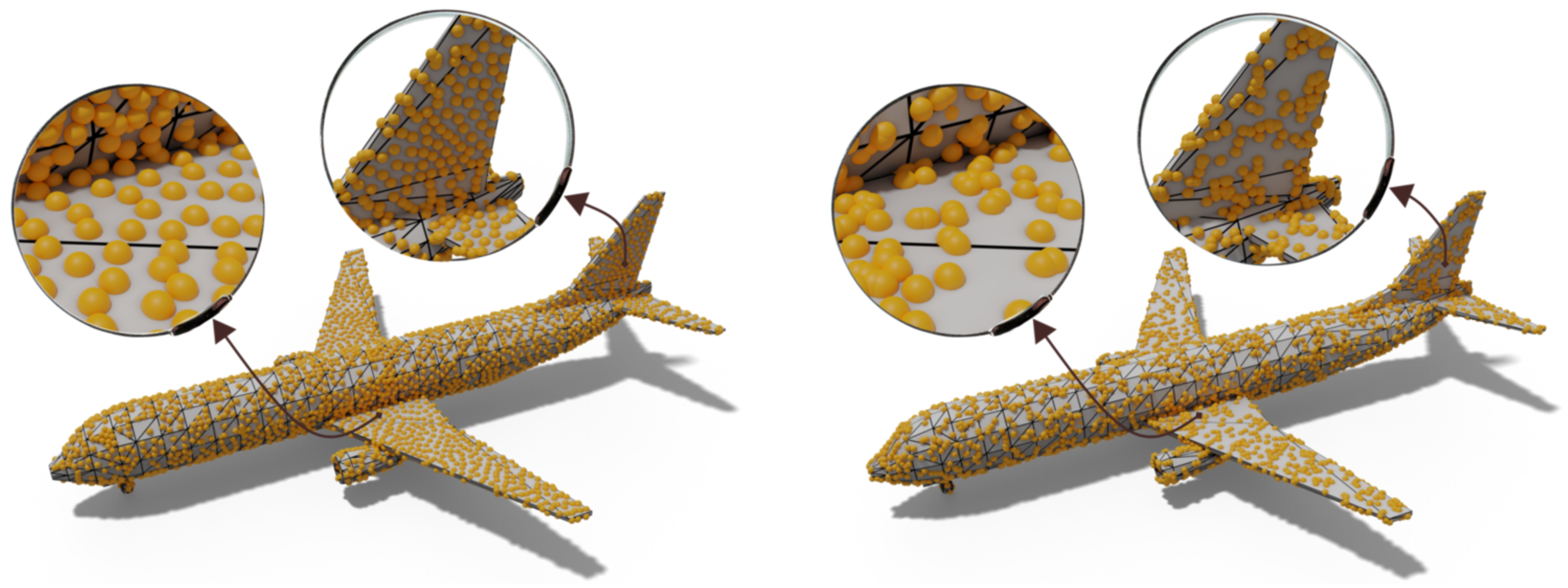}
    \caption{Plane of the ShapeNet dataset sampled with 5k points. \textbf{Left:} Point cloud produced by MongeNet. \textbf{Right:} Point cloud produced by the random uniform sampler. Note the clamping pattern across the mesh produced by the random uniform sampling approach. }
    \label{fig:aviaoLupa}
\end{figure*}

Recently, computer vision researchers have demonstrated an increasing interest in developing deep learning models for 3D data understanding~\cite{Tatarchenko2019,cao2020comprehensive}. As successful applications of those models, we can mention single view object shape reconstruction~\cite{wang2018pixel2mesh,gupta2020neural}, shape and pose estimation~\cite{Kulon2020CVPR}, point cloud completion and approximation~\cite{Hanocka2020p2m}, and brain cortical surface reconstruction~\cite{SantaCruz2020deepcsr}.

A ubiquitous and important component of these 3D deep learning models is the computation of distances between the predicted and ground-truth meshes either for loss computation during training or quality metric calculations for evaluation. 
In essence, the mainstream approach relies on sampling point clouds from any given meshes to be able to compute point-based distances such as the Chamfer distance~\cite{borgefors1984distance} or earth mover's distance~\cite{rubner1998metric}.  
While there exist in-depth discussions about these distance metrics in the literature \cite{fan2017point,liu2020morphing}, the mesh sampling technique itself remains uncharted, where the uniform sampling approach is widely adopted by most of the 3D deep learning models \cite{wang2018pixel2mesh,mescheder2019occupancy,gupta2020neural}.
This approach is computationally efficient and is the only mesh sampling method available in existing 3D deep learning libraries like PyTorch3D~\cite{ravi2020pytorch3d} and Kaolin~\cite{kaolin2019arxiv}. However, it does not approximate the underlying surface accurately since it produces a clustering of points (clamping) along the surface resulting in large under-sampled areas and spurious artifacts for flat surfaces. For instance, Figure~\ref{fig:mainPicturePaper} shows that the Voronoi Tessellation for uniform sampled points is distributed irregularly, while Figure~\ref{fig:aviaoLupa} highlights sampling artifacts in the tail and wings of an airplane from ShapeNet.

In this paper, we revisit the mesh sampling problem and propose a novel algorithm to sample point clouds from triangular meshes. We formulate this problem as an optimal transport problem between simplexes and discrete Dirac measures to compute the optimal solution. Due to the computational challenge of this algorithm, we train a neural network, named MongeNet, to predict its solution efficiently.

As such, MongeNet can be adopted as a mesh sampler during training or testing of 3D deep learning models providing a better representation of the underlying mesh. As shown in~Figure~\ref{fig:mainPicturePaper}, MongeNet sampled points result in uniform Voronoi cell areas which better approximate the underlying mesh surfaces without producing sampling artifacts (also exemplified in Figure~\ref{fig:aviaoLupa}). 

To evaluate our approach, we first compare the proposed sampling technique to existing methods, especially those within the computer graphics community. We show that our sampling scheme better approximates triangles for a reduced number of sampled points. Then, we evaluate the usefulness of MongeNet on 3D deep learning tasks as a mesh approximator to input point clouds using the state-of-the-art Point2Mesh model \cite{Hanocka2020p2m}. To conduct an in-depth analysis of our results, we detail why our method is performing better in the computation of the distance between two point clouds stemming from the same object surface.
In all tasks, the proposed approach is more robust, reliable, and better approximates the target surface when compared to the widely used uniform mesh sampling technique. We demonstrate a significant reduction in approximation error by a factor of 1.89 for results computed on the ShapeNet dataset.


\section{Related work}

\begin{figure*}[!ht]
    \centering
    \includegraphics[width=.24\textwidth]{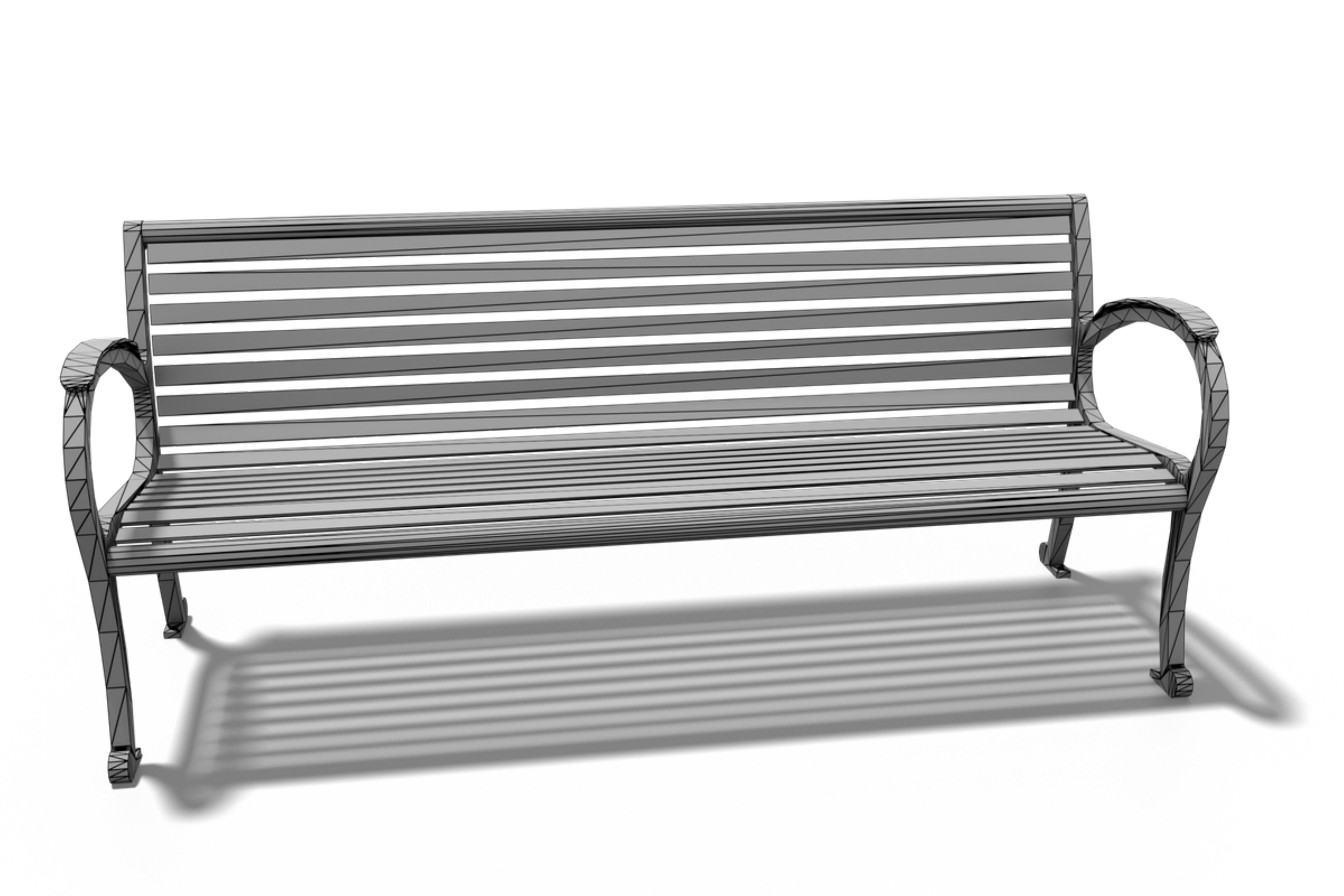}
    \includegraphics[width=.24\textwidth]{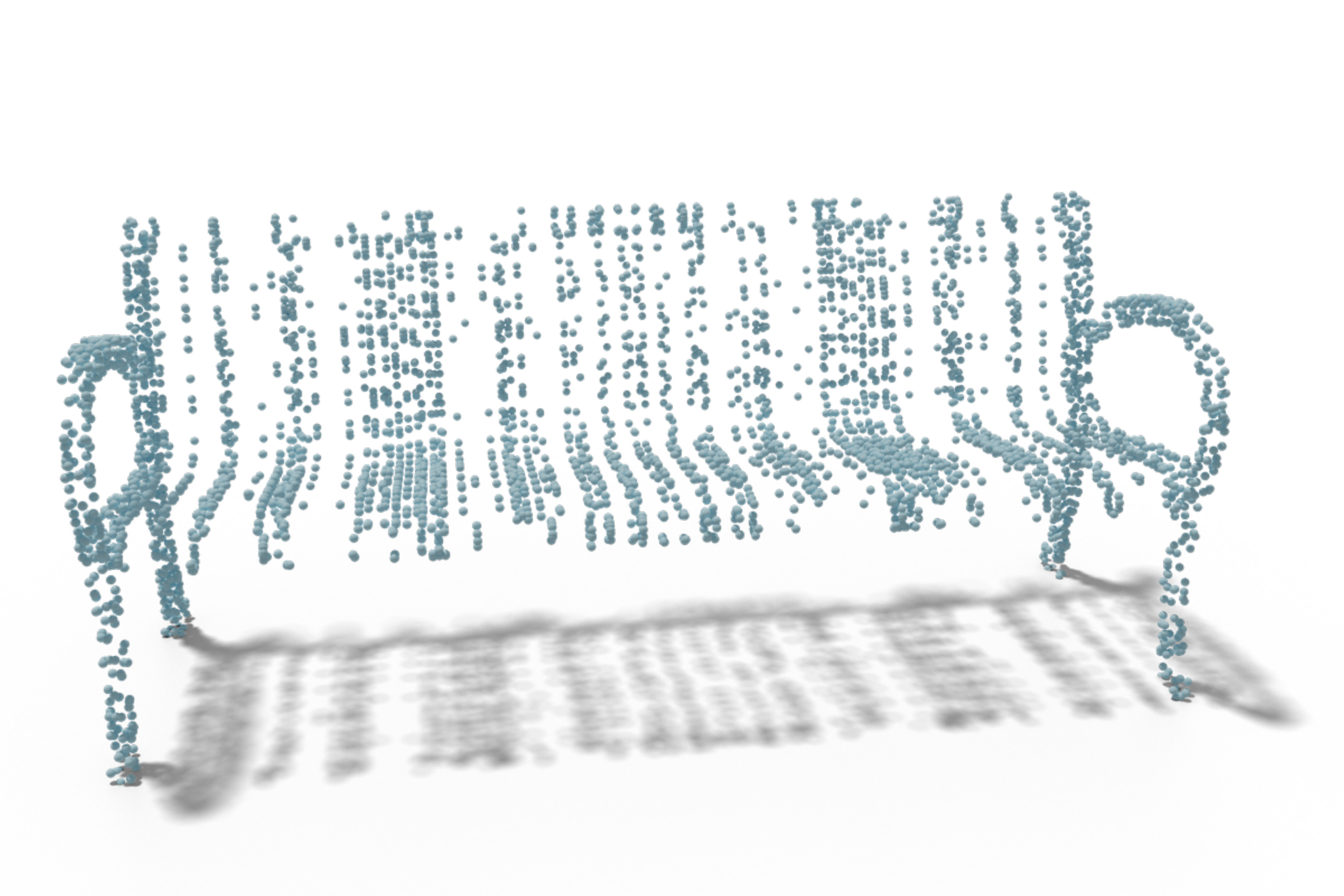}
    \includegraphics[width=.24\textwidth]{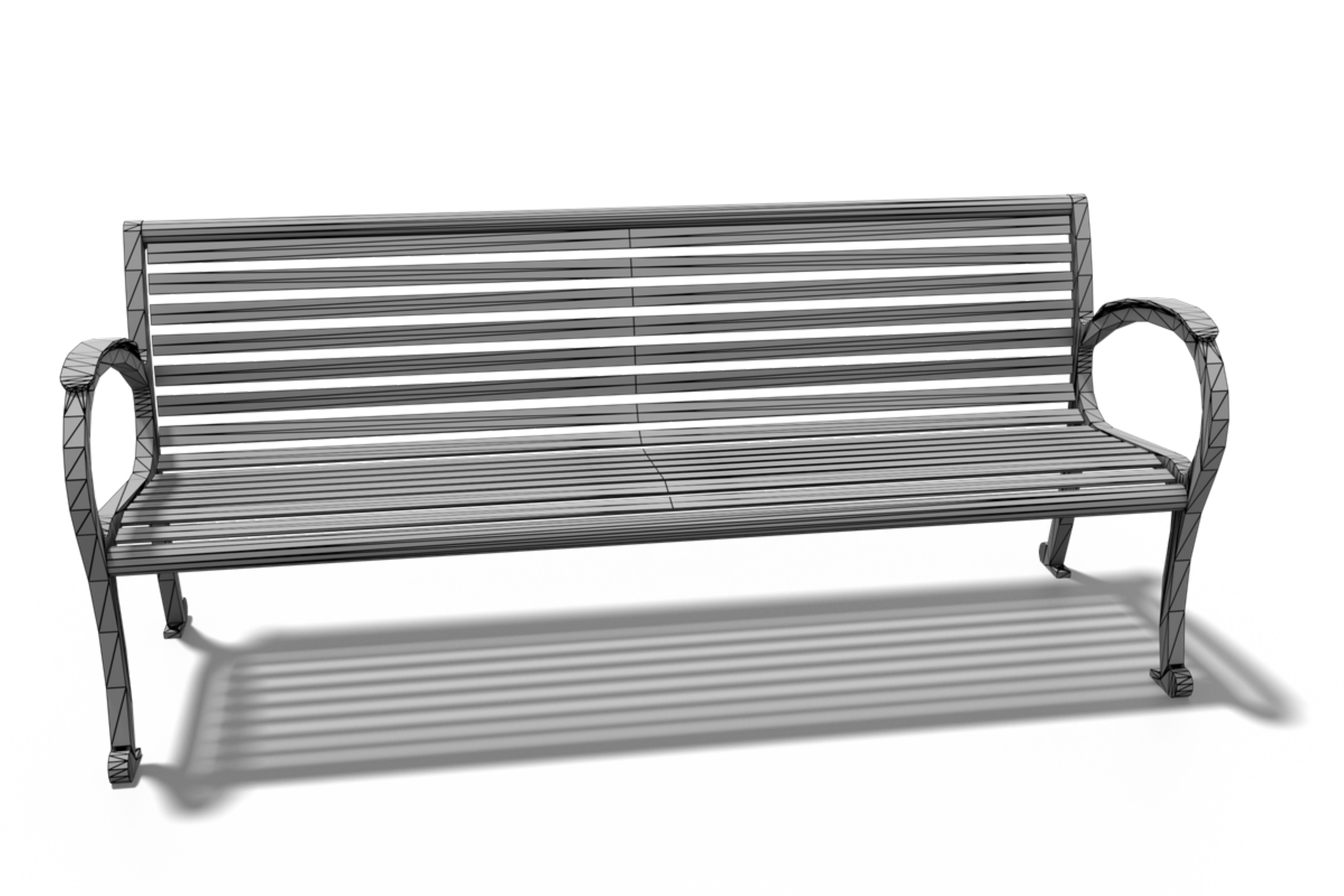}
    \includegraphics[width=.24\textwidth]{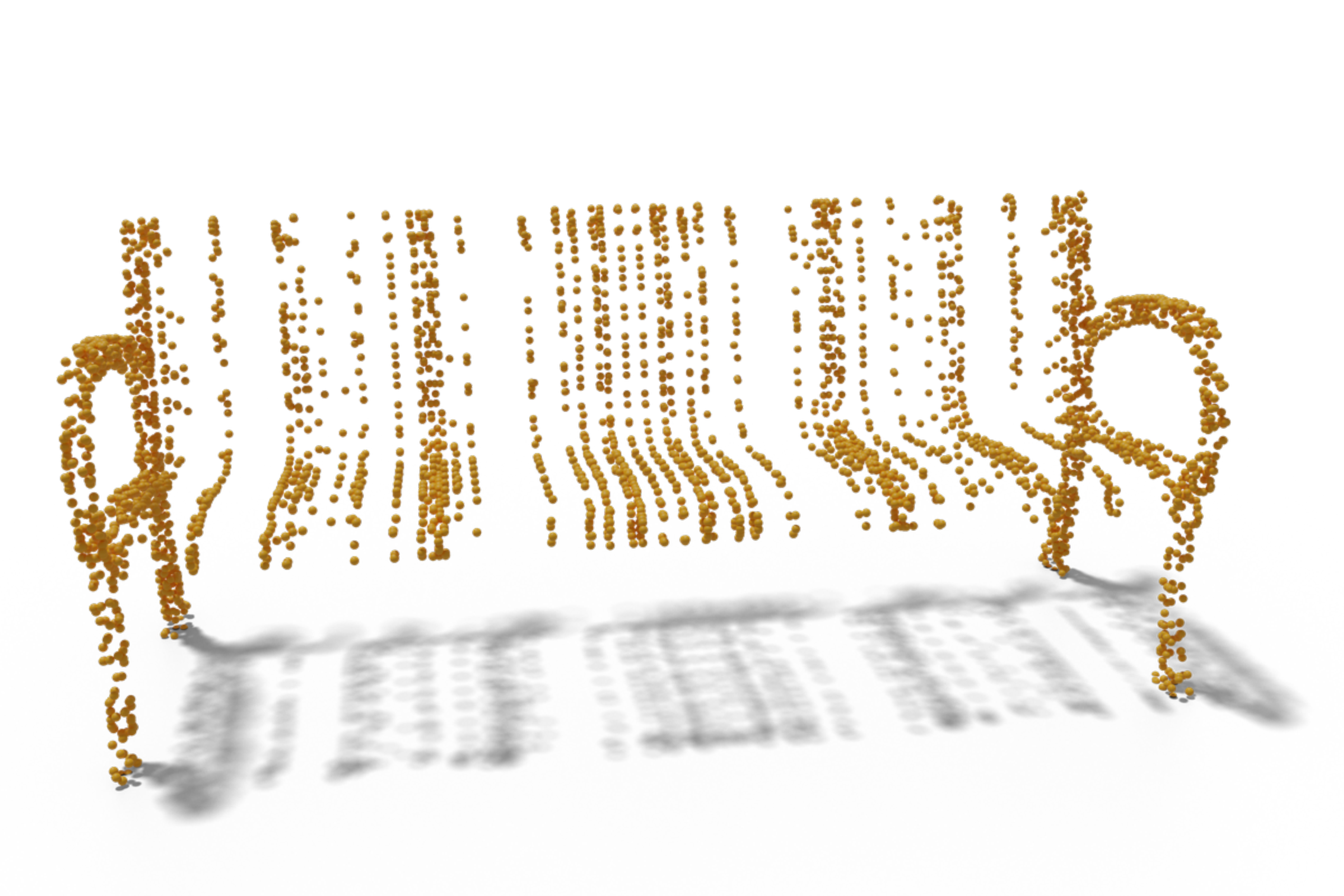}
    \caption{Two point clouds with 3K points drawn with the BNLD~\cite{perrier2018sequences} quasi-random sequence for two different triangulations describing the same 3D model. One would expect to have a very close distance between the point clouds, however, due to the deterministic nature of the sampler, some artefactual patterns appear.}
    \label{fig:SameContinuousObject}
\end{figure*}

Sampling theory has been thoroughly studied for computer graphic applications~\cite{halton1964radical,shirley1991discrepancy,mccool1992hierarchical,kollig2002efficient}, especially for rendering and geometry processing algorithms that rely on Monte-Carlo simulation. Uniform random sampling is subject to clamping, resulting in approximation error and a poor convergence property. Several improvements over the basic uniform sampling technique have been proposed; these rely on techniques that allow the ability to enforce an even distribution of the samples. Poisson disk samplers~\cite{bridson2007fast} reject points that are too close to an existing sample, thereby enforcing a minimal distance between points. Jittering methods~\cite{cook1986stochastic,christensen2018progressive} use a divide and conquer approach to subdivide the sampling space into even-sized strata and then draw a sample in each. Quasi-random sequences~\cite{sobol1967distribution,perrier2018sequences} are deterministic sequences that result in a good distribution of the samples.

All of these samplers are very efficient, allowing the ability to sample hundreds of thousands of points within seconds, and they benefit from good asymptotic properties. However, as will be described in Section~\ref{sec:comparison}, they do not provide an optimal approximation property when the number of samples is limited. This limited number of samples is typical in numerous geometric deep learning applications where less than a hundred points are sampled per triangle.

Sampling a mesh using a point cloud has recently been  proposed with elegant computational geometry algorithms~\cite{merigot2018algorithm,xin2016centroidal,wang2015intrinsic}, generalizing the euclidean Centroidal Voronoi Tessellation (CVT) for surfaces. However, the best approaches have a time complexity in $\mathcal{O}(n^2 \log n)$ which remain computationally non-affordable for computer vision applications. Training a deep learning model requires extensive use of sampling, and faster methods are needed. Indeed for a large number of applications~\cite{wang2018pixel2mesh,hanocka2020point2mesh,gupta2020neural}, this sampling is performed online and at each training step when a new mesh is predicted by the neural network.

To circumvent all the previously described impediments, we propose MongeNet, a new computationally efficient sampling method (0.32 sec for sampling 1 million points on a mesh with 250K faces), while having good approximation properties in the sense of the optimal transport distance between the generated point cloud and the associated mesh.

Related to our work, \cite{lang2020samplenet} propose a point cloud subsampling technique that selects the most relevant points of an input point cloud according to a given task, whereas \cite{liu2020morphing} propose a technique to complete sparse point clouds. Our technique is different since we propose to produce a more accurate and efficient point cloud estimate from a continuous object like a triangular mesh.

\section{Methods}

The quality of mesh sampling can be grasped visually: given a mesh, one expects that a generated point cloud should be distributed evenly over the mesh, describing its geometry accurately as illustrated in Figure~\ref{fig:aviaoLupa}, and uniform random sampling does not meet that expectation. Our proposed sampling is based on optimal transport: given a set of points, how to optimize their positions to minimize the quantity of energy needed to spread their masses on the mesh. Or less formally, given a uniform density of population, how to optimally place click-and-collect points to minimize the overall travel time. We coin our neural network MongeNet in tribute to the mathematician Gaspard Monge and his innovative work on optimal transport.

\subsection{The sampling problem}

To evaluate the distance between two objects, a natural framework is to use an optimal transport distance~\cite{villani2008optimal,santambrogio2015optimal,peyre2019computational}, while representing the mesh objects and point clouds as probability measures. Given a transportation cost $c$, the distance between two probability measures $\mu$ and $\nu$ defined on $X$ (resp. $Y$) is the solution of the following minimization problem,
\begin{equation}\label{eqn:transportOptimalDistance}
\min_{\gamma \in \Pi(\nu,\mu)} \int_{X\times Y} c(x,y) d\gamma(x,y),
\end{equation}
where $\Pi(\nu,\mu)$ is the set of couplings of $(\mu,\nu)$.

In order to consider this type of distance, one has to cast the object of interest into probability measures. For a triangular mesh $T$ of simplexes $(t_i)_{i=1\cdots m}$ one can consider the continuous measure $\mu_c$ carried by a union of simplexes which is defined for any Borel set $B \subseteq \mathbb{R}^3$ as, 

\begin{equation}
\mu_c^T(B) = \frac{1}{| T |} \sum_{t_i \in T} \int_{B \cap t_i} d\mathcal{H}^2(x),
\end{equation}
with $|T|$ the total area of the mesh and  $\mathcal{H}^2$ the 2-dimensional Hausdorff measure. For discrete point clouds with points $(\mathbf{x}_i)_{i=1\cdots n}$ the discrete measure $\mu_d$ reads as a sum of $n$ equi-weighted Dirac masses with mass $\frac 1 n $,
\begin{equation}
\mu_d^T = \frac{1}{n} \sum_{t_i \in T} \sum_{j=1}^{ a_i}  \delta_{\mathbf{x_j}}, {\text{  with  }} \mathbf{x_j}\sim U(t_i),
\end{equation}
where $a_i$ is proportional to the area of the $i$-th triangle $t_i$ with $\sum_i a_i = n$ and where $U(t_i)$ is the uniform distribution on simplex $t_i$. Note that computing an optimal transport distance between two continuous objects is practically intractable for our applications since it requires continuous probability density~\cite{benamou2000computational,papadakis2014optimal}. 

To solve this problem, discretization of a continuous mesh is performed, then, with these point cloud discretizations one can compute a fast distance. More specifically, the number of sampled points per face is drawn according to a multinomial distribution parameterized by the number of queried points $n$. The points on the face are drawn using the square-root parametrization~\cite{turk1990generation,ladicky2017point}, which given the triangle $t_i = (V_i^1,V_i^2,V_i^3)$ reads as,
\begin{equation}
p = (1 - \sqrt{u_1})  V_i^1  + \sqrt{u_i}u_2 V_i^2 + (1-u_2)\sqrt{u_1}V_i^3, \label{eq:squareRootParam}
\end{equation}
where $p \in \mathbb{R}^3$ is a sampled point on the triangle $t_i$ and $u_1,u_2 \sim U([0,1])$.
Providing such a discretization of the triangular mesh allows the use of fast metrics such as the Chamfer distance (CD), earth mover's distance (EMD), and Hausdorff distance. The common issue of this technique is that one has to control the discretization error stemming from converting a continuous measure to a discrete one.
For two meshes $T^1$ and $T^2$, the distance between the two point clouds can be used as a proxy for the distance between the two corresponding meshes. Indeed, the triangle inequality for any Wasserstein metrics $W$ yields,
\begin{align}
&W(\mu_c^{T^1},\mu_c^{T^2}) \leq \nonumber\\
&\underbrace{W(\mu_c^{T^1},\mu_d^{T^1}) + W(\mu_c^{T^2},\mu_d^{T^2})}_{\text{discretization errors}} + W(\mu_d^{T^1},\mu_d^{T^2}). \label{eqn:TriangleInegality}
\end{align}

In this paper, we propose to reduce the discretization error with our new sampling technique to better represent continuous objects with point clouds. Notwithstanding this property, we want to bring the reader's attention towards the re-meshing invariance of the proposed method. From \eqref{eqn:TriangleInegality}, suppose that $T^1$ and $T^2$ are two triangulation instances from the same mesh, obtained with edge splitting  or re-meshing. The distance between the two continuous objects is null and one would like the discretized form of this distance to be null as well. As illustrated in Figure~\ref{fig:SameContinuousObject}, sampled point clouds issued from deterministic quasi-random sequences~\cite{perrier2018sequences} can produce distinctive results. We observed that for very structured meshes, interrelation between points and artefactual patterns could appear, further reducing the accuracy of the estimated distance between them.

\subsection{MongeNet}

MongeNet is a model intended to replace the uniform sampling technique described in Equation~\eqref{eq:squareRootParam}: given the coordinates of a triangle (2-simplex) $t_i=(V_i^1,V_i^2,V_i^3)$ and a number of sampling points $\ell$, one generates a random sequence $S_k$ of points lying on $t_i$.
MongeNet is trained to minimize the 2-Wasserstein distance (optimal transport distance~\eqref{eqn:transportOptimalDistance} for the squared euclidean metric) between the measures carried  by uniform Dirac masses $\mu(S_k)$ with positions $\mathbf{s}_k$ and the continuous measure $\mu(t_i)$ carried by the simplex $t_i$. Without loss of generality, MongeNet approximates the solution of, 
\begin{equation}
    \argmin_{(\mathbf{s}_k)_{k = 1 \cdots \ell} \in \mathbb{R}^{\ell\times3}} W_2^2 (\mu(t_i),\mu(S_k)), \label{eq:ViadoTransport}
\end{equation}
with $W_2$ the 2-Wasserstein distance. Such a problem falls in the semi-discrete optimal transport framework.
In this formalism, one can compute the optimal transport distance solving a convex optimization problem~\cite{merigot2011multiscale,de2012blue,levy2015numerical}. This problem relies on a powerful geometrical tool, the Laguerre Tesselation or power-diagram, corresponding to a weighted version of the Voronoi tessellation. To find a sampling with good quantization noise power~\cite{levy2015numerical,du1999centroidal} there exist iterative algorithms \cite{merigot2011multiscale}. The resulting point cloud on a triangle is a Centroidal Voronoi Tessellation (CVT) such that each Dirac mass is placed exactly on the center of mass of its Voronoi cell~\cite{aurenhammer1992minkowski}.
Solving this problem is time consuming and computationally too expensive for most applications. Instead, we propose to learn the optimal points position using a deep neural network so that a point cloud with a good approximation property can be generated within a few centiseconds. Such an optimal point sampling learned by MongeNet is displayed in Figure~\ref{fig:mainPicturePaper}~and~\ref{fig:aviaoLupa}.

\textbf{Network Training.} MongeNet is a feed-forward neural network denoted as $f_\theta$, and parameterized by its learnable parameter $\theta$. It takes as input a triangle $t$, a discrete number of output points $\ell \in \llbracket 1, 30 \rrbracket$, and a random noise $p\in \mathbb{R}$ randomized following a standard normal distribution. It outputs a random sequence of $\ell$ points $f_\theta(t,\ell,p) \in \mathbb{R}^{3\times\ell}$.

As a supervised model, MongeNet is trained to minimize the regularized empirical risk across a given training set $\mathcal{D} = \{(t_i, S_{i})_{i=1..N}\}$ consisting in a sequence of pairs of triangles and sampled points,
\[
\argmin_{\theta} \sum_{i=1}^{N} \sum_{\ell=1}^{30} \mathcal{L}(f_\theta(t_i,\ell,p_i),S_{i}),
\]
with  $p_i \sim \mathcal{N}(0,1)$ and the loss function $\mathcal{L}$ defined as,
\begin{align*}
\mathcal{L}(t,\ell,p,S) &= W_2^\varepsilon(f_\theta(a,\ell,p),S) \nonumber\\
& -\alpha W_2^\varepsilon(f_\theta(a,\ell,p),f_\theta(a,\ell,p')).
\end{align*}
where $W_2^\varepsilon$ is the $\varepsilon$-regularized optimal transport~\cite{cuturi2013sinkhorn,charlier2020kernel}, and $p' \sim \mathcal{N}(0,1)$ is an adversarial Gaussian noise input that encourages entropy (different point locations) in the predicted point cloud.
We minimize this loss using the Adam Optimizer with batches of 32 triangles for approximately 15K iterations until the validation loss reaches a plateau. We set the blending parameter to $\alpha=0.01$ and the Entropic regularization $\varepsilon = 5\cdot 10^{-5}$.

\textbf{Generation of the dataset.}
The training examples consist of $19,663$ triangles sampled with $30, 50, 100, 200, 300, 500, 1000,$ and $2000$ points. The optimal point configurations are obtained using blue-noise algorithms provided in~\cite{de2019differentiation,lebrat2019optimal}. In order to obtain a continuous measure from the triangle coordinates, we discretize the measure carried by each triangle using $\mathcal{Q}^1$ quadrilateral finite elements on a grid of resolution $500\times 500$.

\textbf{Generating an arbitrary number of points per face.} The current implementation of MongeNet allows the computation of up to 30 points per-face. This is generally sufficient for most deep-learning applications. However, since this number can be arbitrarily large, we provide a local refinement technique that allows sampling any number of points per face. It amounts to splitting the largest edge of the triangle containing more than 30 points, to obtain two smaller triangles with a reduced area. This splitting operation is repeated until all of the faces are sampled with at most 30 points.

\begin{figure}
    \centering
    \includegraphics[width=\linewidth]{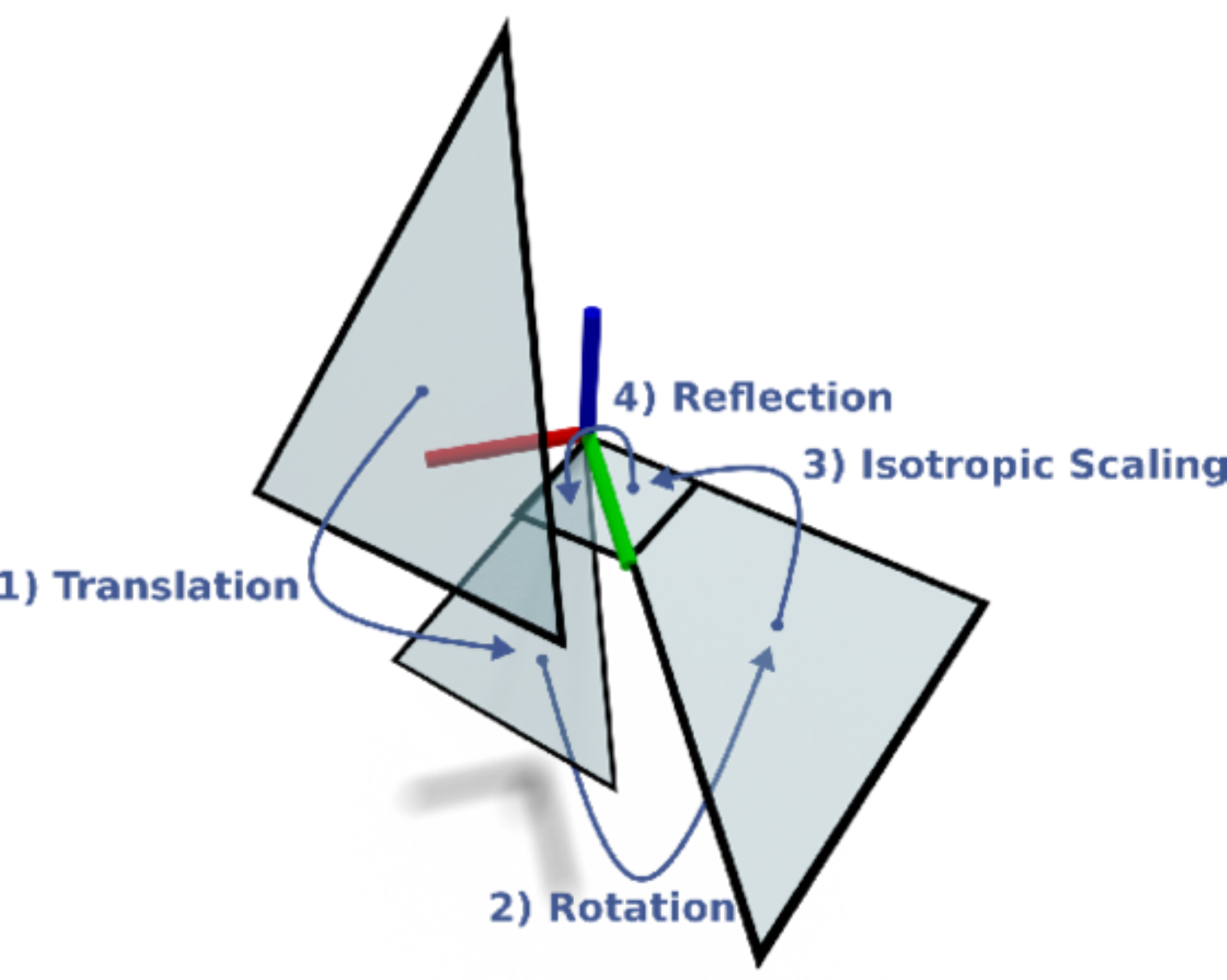}
    \caption{The query face is mapped towards the unit square using translation, rotation and reflection (isometries) and a uniform scaling (similarity).}
    \label{fig:mappingUnitSquare}
\end{figure}

\textbf{Reduction to invariant learning-problem.} The shape space for triangles is a two-dimensional manifold~\cite{kendall2009shape}, thence up to normalization, any triangle can be parameterized with 2 unknowns. We thus map all the triangles to the unit square $[0,1]^2$, fixing the longest edge coordinates's to $(0,0)\rightarrow(0,1)$. This operation is depicted in Figure~\ref{fig:mappingUnitSquare}. All the operations are angle preserving so that the inverse transformations will conserve geometric optimality of the point cloud predicted on the unit square.

\textbf{Network Architecture.} Given an invariant triangle representation and a random variable sampled from a standard Gaussian distribution, MongeNet outputs 30 sets ranging from 1 to 30 random points. For computational efficiency, these output sets are tensorized as a matrix with 465 rows and 2 columns such that the $i$-th sampled point in the $j$-th output set with $j$ samples is in the $\frac{j\left( j - 1\right)}{2} + i - 1$ row of this matrix. The network architecture consists of three linear layers of 64 output neurons with a ReLU activation function and Dropout between layers. The output uses sigmoid functions to constrain the prediction to a unit square which is remapped to points in the input triangle using the area-preserving parameterization with low-distortion presented in \cite{heitz2019low}. It is then remapped to the original triangle using the inverse of the transformation described in Figure~\ref{fig:mappingUnitSquare}.

\textbf{Complexity analysis.}
The complexity of MongeNet is tantamount to the evaluation of a feedforward neural network. It is done rapidly on GPU and the computation across triangles is performed by batch. MongeNet's performance ({\bf MN}) is competitive with pytorch3D's random uniform sampler ({\bf RUS}), we compare their runtime in Table~\ref{tab:comparisonSpeed}.
\begin{table}[h]
\begin{adjustbox}{width=.95\linewidth}
    \begin{tabular}{l|c|c|c|c|c|c}
         \# Faces & 10k & 20k & 30k & 40k & 60k & 80k \\
         \hline
         {\bf RUS} & 1.14 ms & 1.50ms & 1.53ms & 1.52ms & 1.53ms & 1.53ms\\
         \hline
         {\bf MN} & 2.89 ms & 5.41 ms & 7.90 ms & 10.5 ms & 16.0 ms& 21.7 ms\\
    \end{tabular}
\end{adjustbox}
\caption{{Runtimes for sampling 20k points on a mesh with an increasing number of faces and using an Nvidia RTX 3090.}}\label{tab:comparisonSpeed}
\end{table}

MongeNet's complexity scales linearly with the number of faces to be sampled. Non-deterministic CPU-based methods described in the next section cannot be used during training due to their longer runtime for sampling a mesh with more that 10k faces ($>1s$).

\section{Experiments}
\subsection{Comparison with pre-existing sampling methods}
\label{sec:comparison}

\begin{figure}[!ht]
    \centering
    \includegraphics[width=\linewidth]{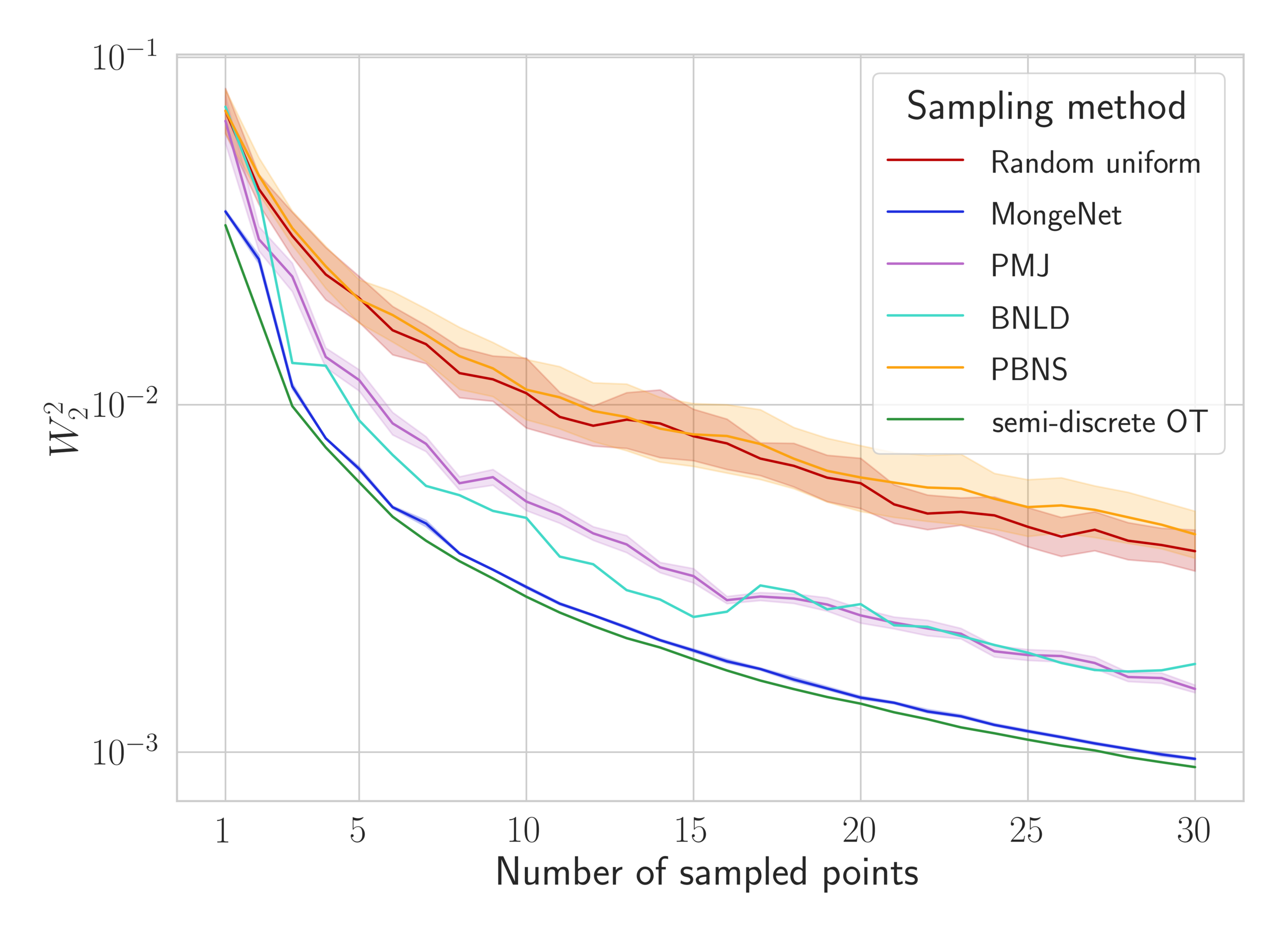}
    \caption{Comparison of different sampling techniques for approximating a triangle with an increasing number of points.}
    \label{fig:ComparisonMethods}
\end{figure}

\begin{figure*}[htb!]
    \centering
    \includegraphics[width=.32\textwidth]{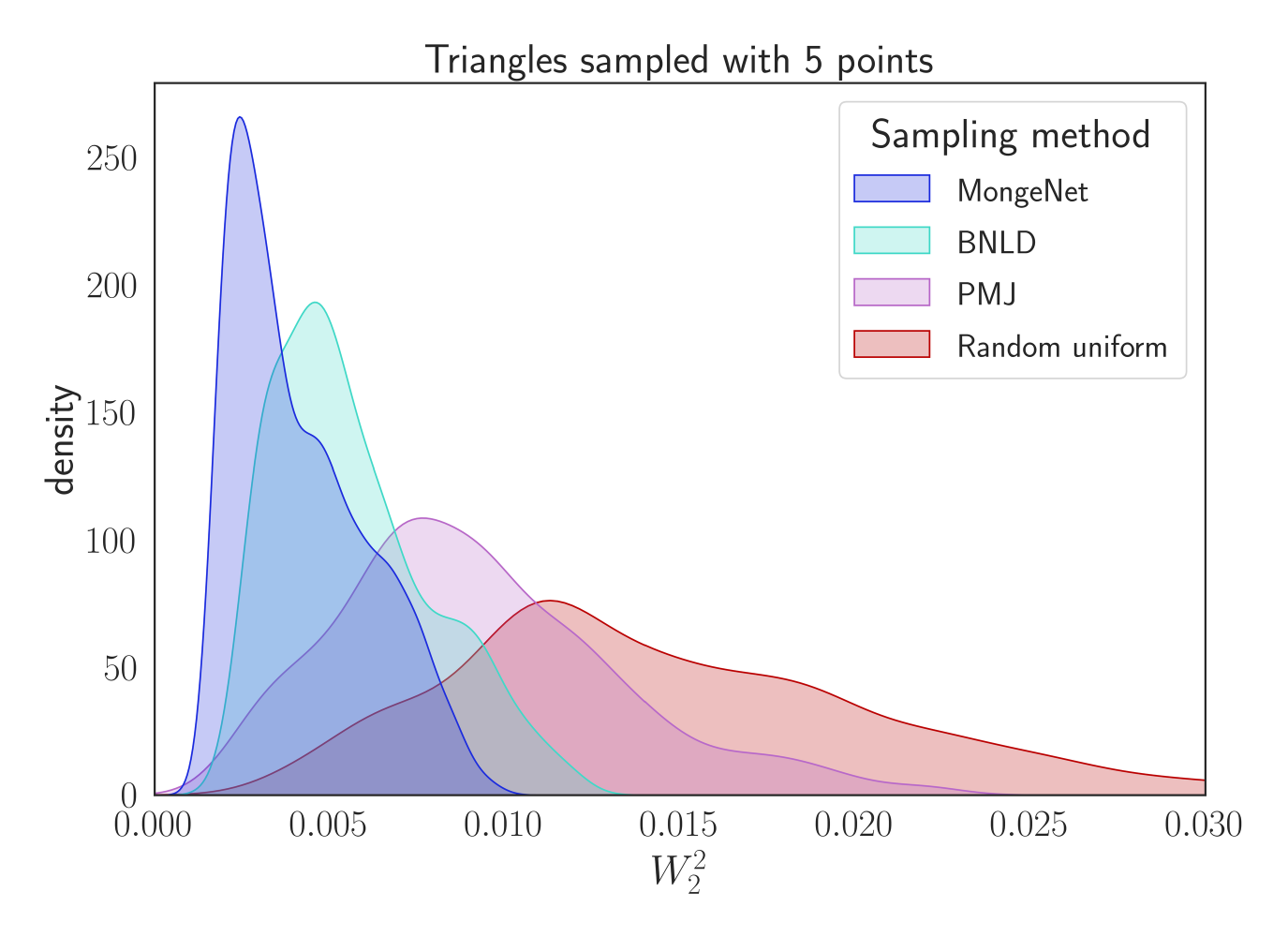}
    \includegraphics[width=.32\textwidth]{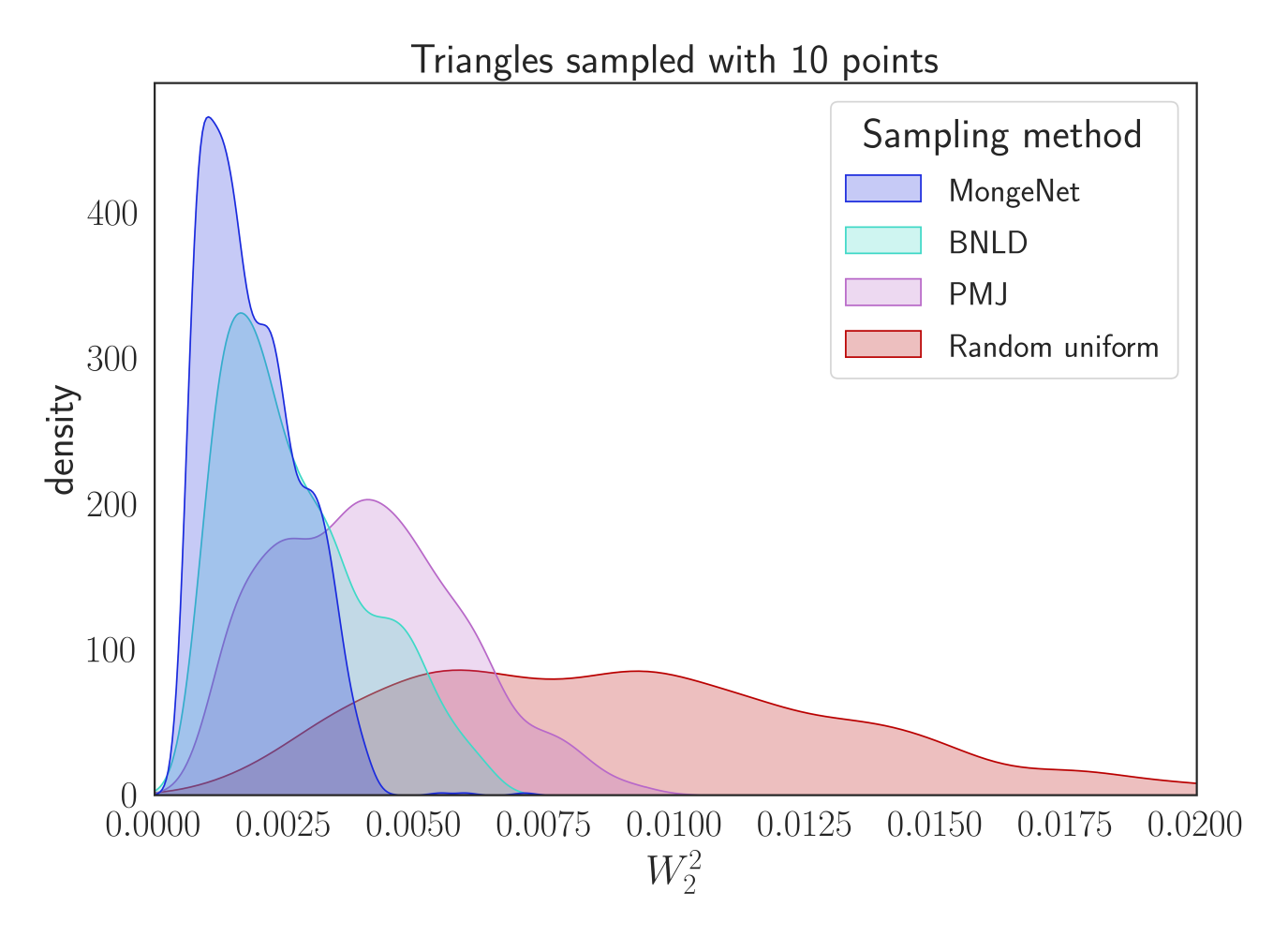}
    \includegraphics[width=.32\textwidth]{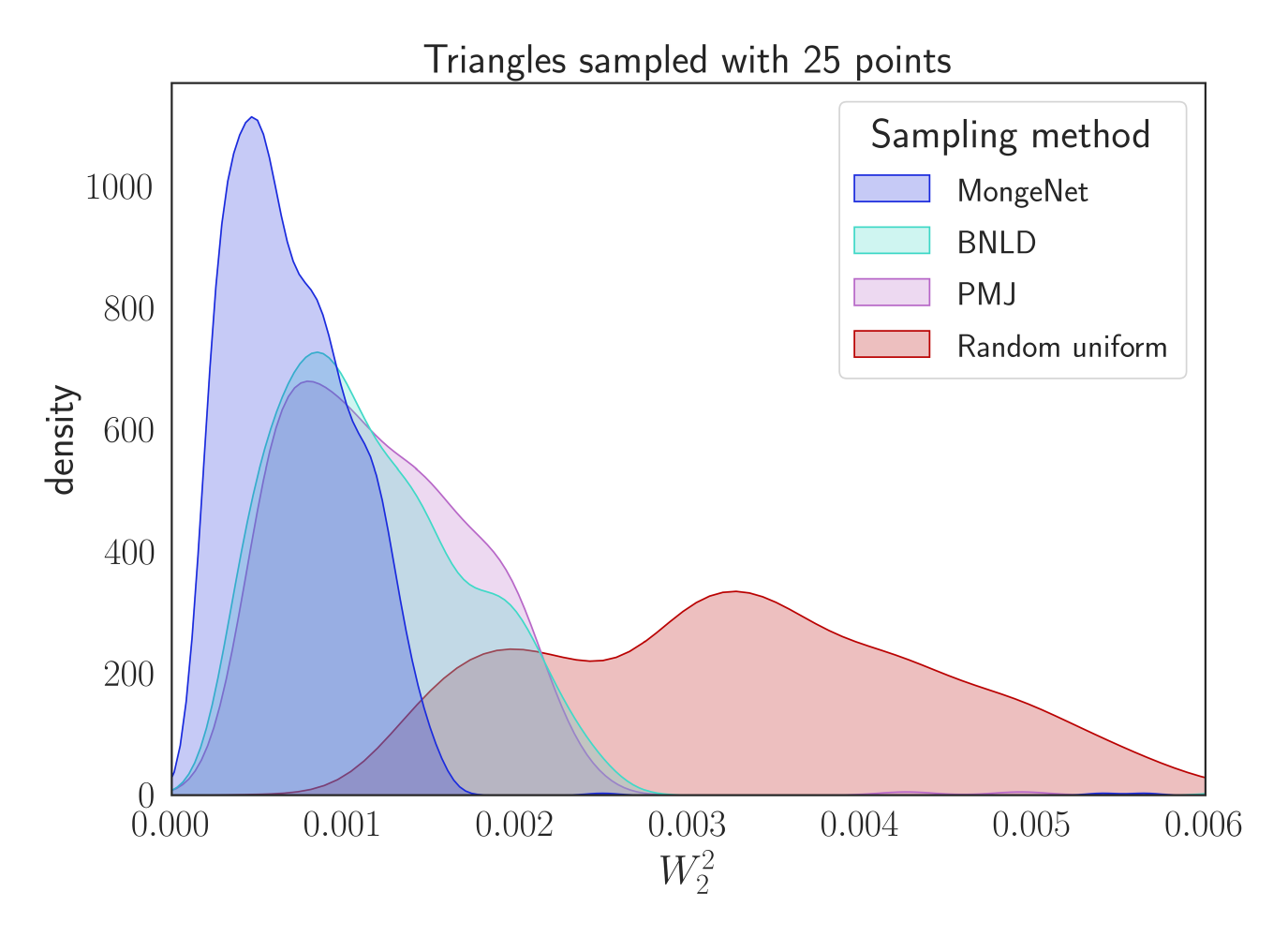}
    \caption{Distribution of the approximation error for 500 triangles drawn uniformly random. From left to right: an approximating point cloud with 5 sampled points, 10 sampled points, and 25 sampled points.}
    \label{fig:triangleSampledUniformly}
\end{figure*}

To evaluate the performance of our method, we compare it to three recent sampling methods: 1) Projective Blue-Noise Sampling (PNBS)~\cite{reinert2016projective} and the dart-throwing algorithm with a rule of rejection that depends on full-dimensional space and projections on lower-dimensional subspaces; 2) progressive multi-jittered sample sequence (PMJ)~\cite{christensen2018progressive}; and 3) Blue-Noise low discrepancies sequences (BNLD)~\cite{perrier2018sequences}, a quasi-random sequence with increased blue noise properties. We compare these methods against the random uniform sampling, and the sampling generated by our model MongeNet for an isosceles triangle and for an increasing number of points and 25 randomly drawn samples. All the distances between the estimated point cloud samples and the continuous object are obtained via computing the distance between the predicted points and $2000$ points sampled on the triangle using a semi-discrete optimal transport blue-noise algorithm~\cite{de2012blue}. The results are reported in Figure~\ref{fig:ComparisonMethods}. 

Since our method approximates the point cloud location in the sense of optimal transport, it exhibits the best approximation error for this metric. The average 2-Wasserstein distance was 1.89 times smaller than the random uniform sampling, 1.30 times smaller than the PMJ sampling method, and 1.25 times smaller for BNLD. The MongeNet samples are different with each random sampling of the latent variables $p$, but the resulting point clouds have a similar error of approximation. Note that due to the deterministic nature of the BNLD sequence, there is no variance in the approximation error.
Of note, we are interested in the sampling patterns for a reduced number of points (typically $<100$) whereas related works ~\cite{reinert2016projective,christensen2018progressive,perrier2018sequences} have focused on sampling patterns for many more points ($>10,000$).

To show the behavior for a larger variety of triangles, we sample randomly 500 triangles uniformly into the unit square and examine the distribution of the approximation error for random uniform sampling, MongeNet, and its two forerunners PMJ and BNLD. We repeat the sampling three times using different random seeds to average the sampling performance across a given triangle. This experiment is summarized for 5, 10, and 25 sampled points in Figure~\ref{fig:triangleSampledUniformly}. Among all of the samplers, MongeNet proposes a triangle sampling with on average the best approximation quality.

\subsection{Approximation error quantification}

\begin{figure*}
    \centering
    \includegraphics[width=\textwidth]{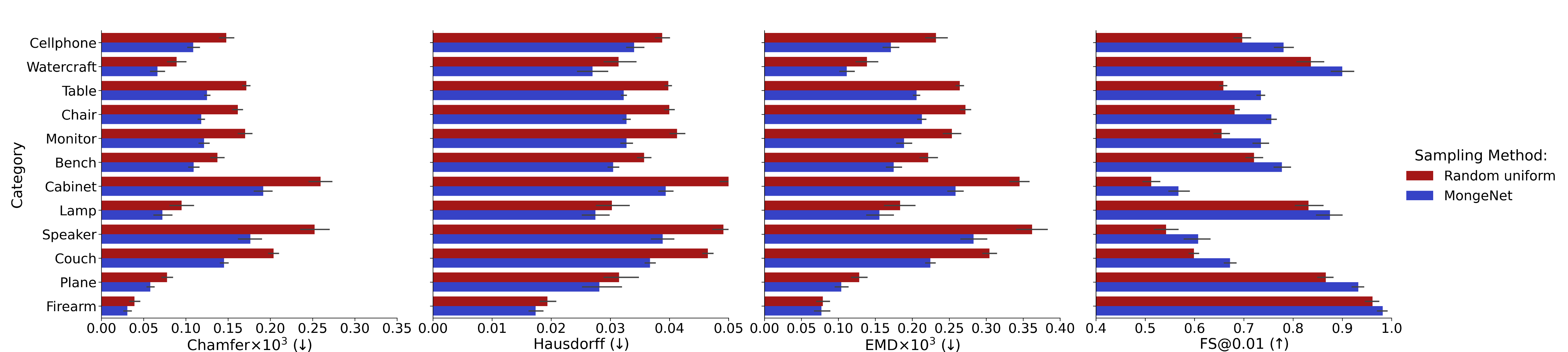}
    \caption{Average approximation errors between two point clouds sampled from different tessellations of the same underlying surface.}
    \label{fig:MetricAcross}
\end{figure*}

\begin{figure}
    \centering
    \includegraphics[width=\linewidth]{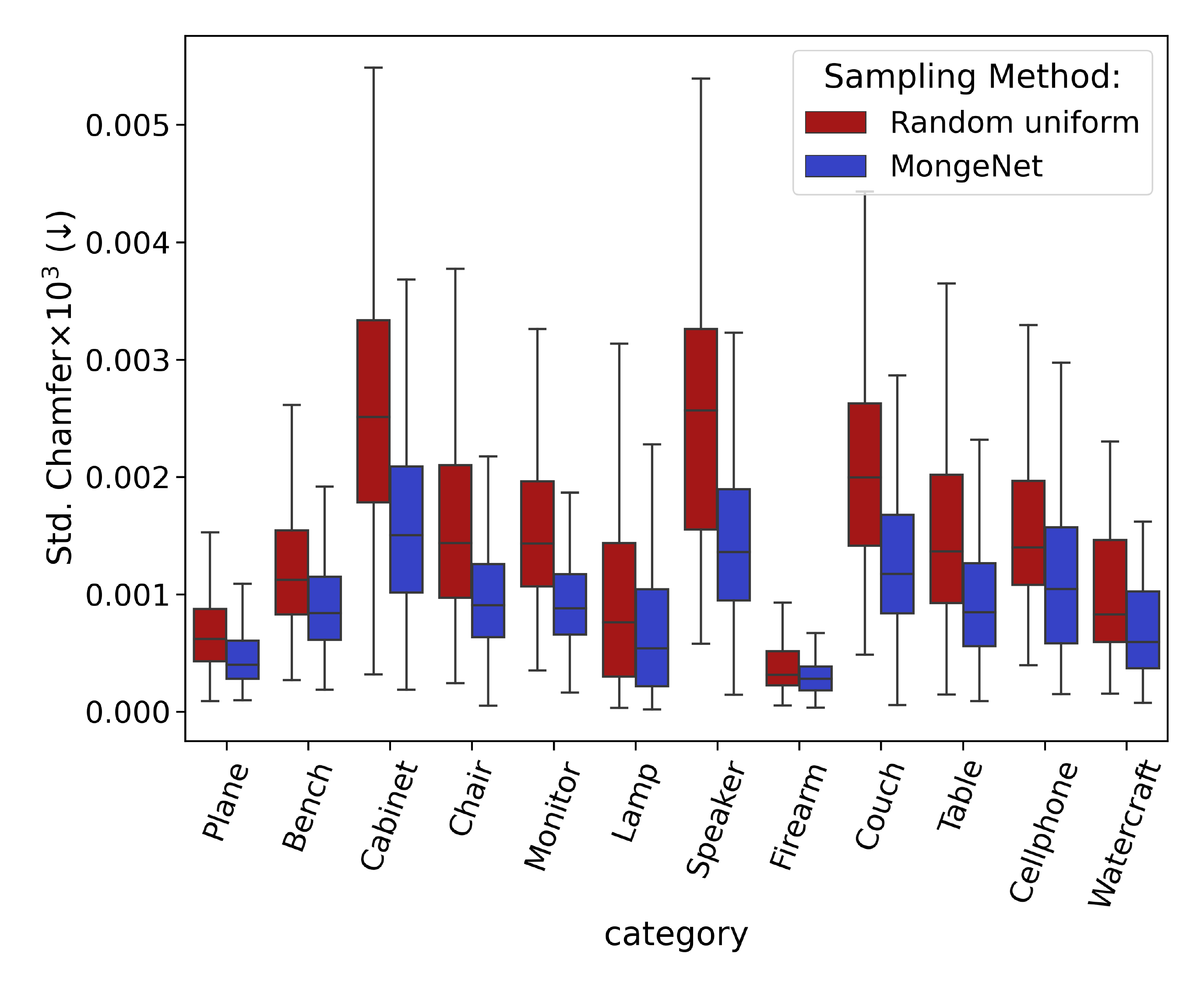}
    \caption{Boxplot of the standard deviation of the approximation error across repetitions in terms of Chamfer distance and grouped by ShapeNet categories.}
    \label{fig:MetricBoxPlot}
\end{figure}

We compare the approximation errors generated by sampling a similar surface from the ShapeNet dataset with MongeNet and a random uniform sampler.
Given a 3D model from a subset of 3200 models from the Shapenet dataset\footnote{The list of models used for this experiment is made available at \href{https://github.com/lebrat/MongeNet/blob/9a54c94c473d7ada5f8499232758915b9d84d067/resources/meshList.txt}{github.com/lebrat/MongeNet/resources/meshList.txt}}, we first sample 100K points using the random uniform sampler. Then we re-mesh randomly 30\% of the faces of the mesh to change its simplexes. With this new remeshing we produce a point cloud of 10K points using either the random uniform sampler or Mongenet. We report the Chamfer distance, the Hausdorff distance, and the F0.01 Score. We also report the EMD distance, but due to hardware limitations, this computation is performed using only 25K points of the original model.
In order to obtain more stable estimates, we reproduce the experiments for 10 repetitions.
In Figure~\ref{fig:MetricAcross} we display the average error produced in function of the sampler method used. Points that are sampled using MongeNet are consistently yielding a lower approximation error. In addition, the variance of the sampling is decreased, as depicted in Figure~\ref{fig:MetricBoxPlot}.

\subsection{Mesh Reconstruction}
\label{sec:Point2Mesh}
We now evaluate MongeNet on reconstructing watertight mesh surfaces to noisy point clouds which is a prerequisite for downstream applications such as rendering, collision avoidance, and human-computer interaction. 
More specifically, we use the Point2Mesh \cite{Hanocka2020p2m} framework as the backbone. It consists of a MeshCNN that iteratively deforms a template mesh to tightly wrap a noisy target point cloud. At each iteration, it minimizes point cloud based distances between a point cloud sampled from the template mesh and the target point cloud. We contrast the performance of this method when configured with the standard uniform sampling and when configured with MongeNet. 
Point2Mesh hyperparameters and the target noisy point clouds used in this experiment are kindly provided by the Point2Mesh authors\footnote{\href{https://github.com/ranahanocka/point2mesh}{https://github.com/ranahanocka/point2mesh}}. 
Since Point2Mesh is an optimization based approach relying on random initialization, we repeat the experiment 10 times and report the average and standard deviation of the aforementioned metrics between the target point cloud and a point cloud sampled with 200K points from the produced final watertight mesh (using the same uniform sampling technique for both methods). Table~\ref{tab:point2mesh} presents the results of this experiment.

\begin{table*}[th!]
\resizebox{\textwidth}{!}{\begin{tabular}{l|ccccc|ccccc|}
 & \multicolumn{5}{c|}{Point2Mesh with \underline{\textbf{MongeNet} sampling}} & \multicolumn{5}{c|}{Point2Mesh with \underline{\textbf{uniform} sampling}}\\
 Objects & CHD$\times10^3$~($\downarrow$) & EMD$\times10^3$~($\downarrow$) & HD~($\downarrow$) & NC~($\uparrow$) & FS@0.01~($\uparrow$) & CHD$\times10^3$~($\downarrow$) & EMD$\times10^3$~($\downarrow$) & HD~($\downarrow$) & NC~($\uparrow$) & FS@0.01~($\uparrow$) \\
\hline
Bull & \makecell{$0.095$ \\ $(\pm 0.054)$} & \makecell{$0.153$ \\ $(\pm 0.116)$} & \makecell{$0.032$ \\ $(\pm 0.009)$} & \makecell{$0.954$ \\ $(\pm 0.010)$} & \makecell{$0.879$ \\ $(\pm 0.121)$} & \makecell{$0.306$ \\ $(\pm 0.345)$} & \makecell{$0.641$ \\ $(\pm 0.885)$} & \makecell{$0.050$ \\ $(\pm 0.022)$} & \makecell{$0.929$ \\ $(\pm 0.037)$} & \makecell{$0.693$ \\ $(\pm 0.287)$} \\
Giraffe & \makecell{$0.637$ \\ $(\pm 0.752)$} & \makecell{$1.046$ \\ $(\pm 1.238)$} & \makecell{$0.053$ \\ $(\pm 0.028)$} & \makecell{$0.880$ \\ $(\pm 0.106)$} & \makecell{$0.675$ \\ $(\pm 0.361)$} & \makecell{$1.161$ \\ $(\pm 1.770)$} & \makecell{$1.903$ \\ $(\pm 3.002)$} & \makecell{$0.061$ \\ $(\pm 0.042)$} & \makecell{$0.863$ \\ $(\pm 0.125)$} & \makecell{$0.638$ \\ $(\pm 0.366)$} \\
Guitar & \makecell{$0.051$ \\ $(\pm 0.007)$} & \makecell{$0.066$ \\ $(\pm 0.010)$} & \makecell{$0.017$ \\ $(\pm 0.001)$} & \makecell{$0.981$ \\ $(\pm 0.002)$} & \makecell{$0.948$ \\ $(\pm 0.030)$} & \makecell{$0.056$ \\ $(\pm 0.008)$} & \makecell{$0.075$ \\ $(\pm 0.007)$} & \makecell{$0.018$ \\ $(\pm 0.001)$} & \makecell{$0.981$ \\ $(\pm 0.001)$} & \makecell{$0.932$ \\ $(\pm 0.027)$} \\
Tiki & \makecell{$0.096$ \\ $(\pm 0.012)$} & \makecell{$0.106$ \\ $(\pm 0.007)$} & \makecell{$0.028$ \\ $(\pm 0.002)$} & \makecell{$0.978$ \\ $(\pm 0.001)$} & \makecell{$0.854$ \\ $(\pm 0.031)$} & \makecell{$0.100$ \\ $(\pm 0.010)$} & \makecell{$0.103$ \\ $(\pm 0.002)$} & \makecell{$0.031$ \\ $(\pm 0.004)$} & \makecell{$0.977$ \\ $(\pm 0.002)$} & \makecell{$0.843$ \\ $(\pm 0.030)$} \\
Triceratops & \makecell{$0.074$ \\ $(\pm 0.024)$} & \makecell{$0.065$ \\ $(\pm 0.017)$} & \makecell{$0.021$ \\ $(\pm 0.004)$} & \makecell{$0.974$ \\ $(\pm 0.004)$} & \makecell{$0.932$ \\ $(\pm 0.058)$} & \makecell{$0.074$ \\ $(\pm 0.037)$} & \makecell{$0.066$ \\ $(\pm 0.023)$} & \makecell{$0.022$ \\ $(\pm 0.007)$} & \makecell{$0.975$ \\ $(\pm 0.004)$} & \makecell{$0.922$ \\ $(\pm 0.092)$} \\
\hline
Overall & \makecell{$0.191$ \\ $(\pm 0.170)$} & \makecell{$0.287$ \\ $(\pm 0.278)$} & \makecell{$0.030$ \\ $(\pm 0.009)$} & \makecell{$0.953$ \\ $(\pm 0.024)$} & \makecell{$0.858$ \\ $(\pm 0.120)$} & \makecell{$0.339$ \\ $(\pm 0.434)$} & \makecell{$0.558$ \\ $(\pm 0.784)$} & \makecell{$0.036$ \\ $(\pm 0.015)$} & \makecell{$0.945$ \\ $(\pm 0.034)$} & \makecell{$0.806$ \\ $(\pm 0.160)$} \\
\hline
\end{tabular}}
\caption{\label{tab:point2mesh} Watertight mesh reconstruction for noisy point clouds using the Point2Mesh framework configured with uniform sampling and MongeNet. The evaluation metrics are computed between the target point cloud and a point cloud sampled with 200K points from the final reconstructed mesh. We report the average and standard deviation of these metrics over 10 repetitions. 
}
\end{table*}

\begin{figure}
    \centering
    \includegraphics[width=\linewidth]{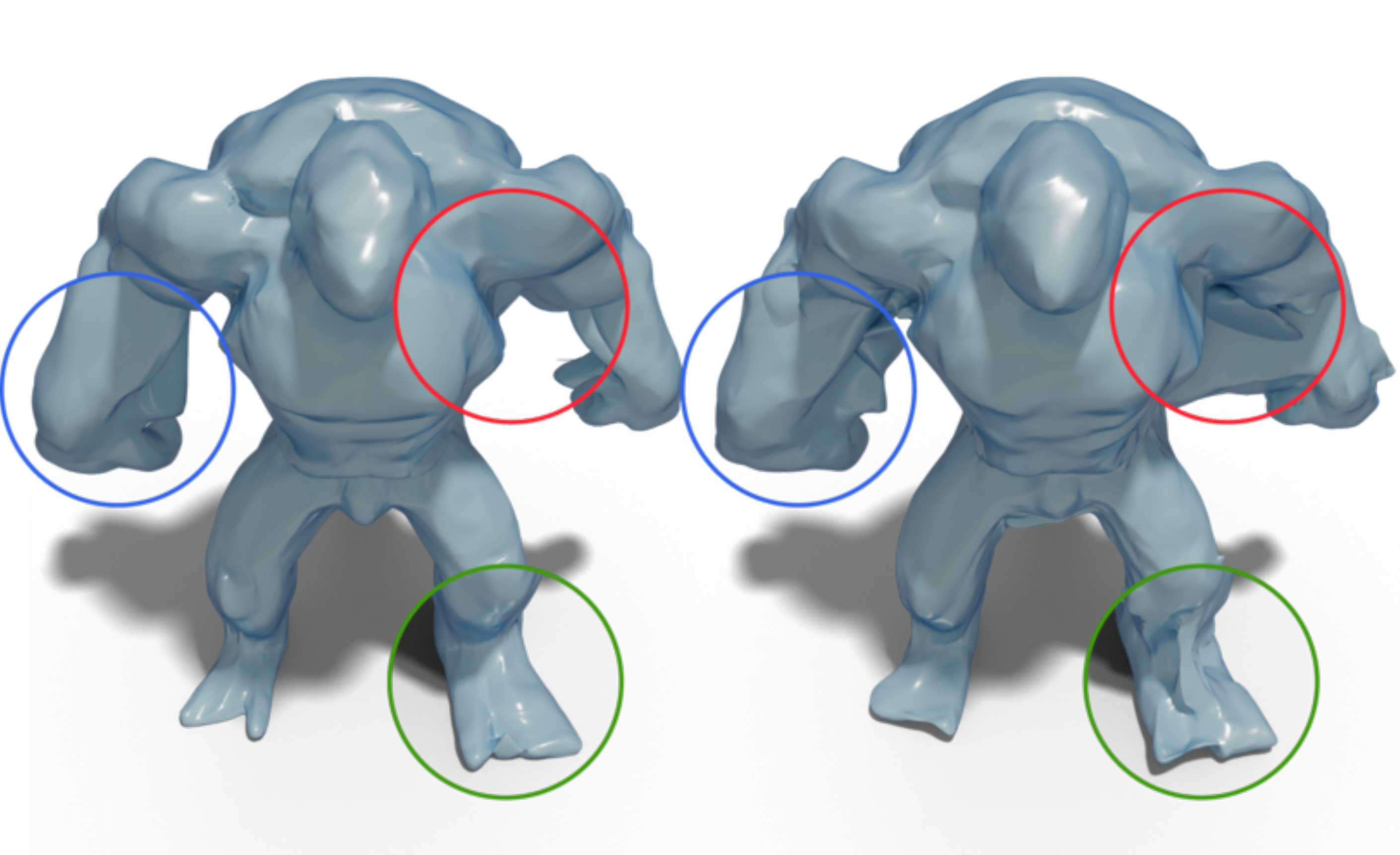}
    \caption{More difficult shape from Thingi10k reconstructed by the Point2Mesh backbone. \textbf{Left:} Point2Mesh with MongeNet sampling. \textbf{Right:} Point2Mesh with random uniform sampling. 
    }
    \label{fig:ComparisonTHingi}
\end{figure}

Point2Mesh equipped with MongeNet sampling outperforms systematically the alternative technique: it improves the Chamfer distance by a factor 1.77, the earth mover's distance by a factor 1.94, and increased on average the FS-0.01 score by 0.052.
We also noticed that larger differences could be observed for more detailed input shapes. We investigated this observation by increasing the difficulty of the reconstruction task using much more complex shapes from the Thingi10k dataset~\cite{zhou2016thingi10k}. As illustrated in Figure~\ref{fig:ComparisonTHingi}, the reconstruction errors are located either in the parts of the mesh with lots of fine details or in the areas with high curvature.

\begin{figure*}[!ht]
\includegraphics[width=\textwidth]{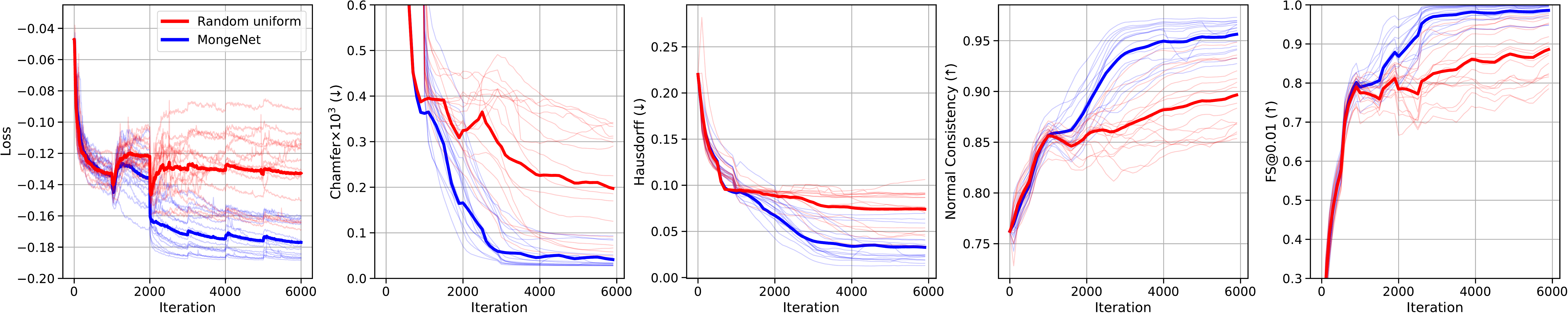}
\includegraphics[width=\textwidth]{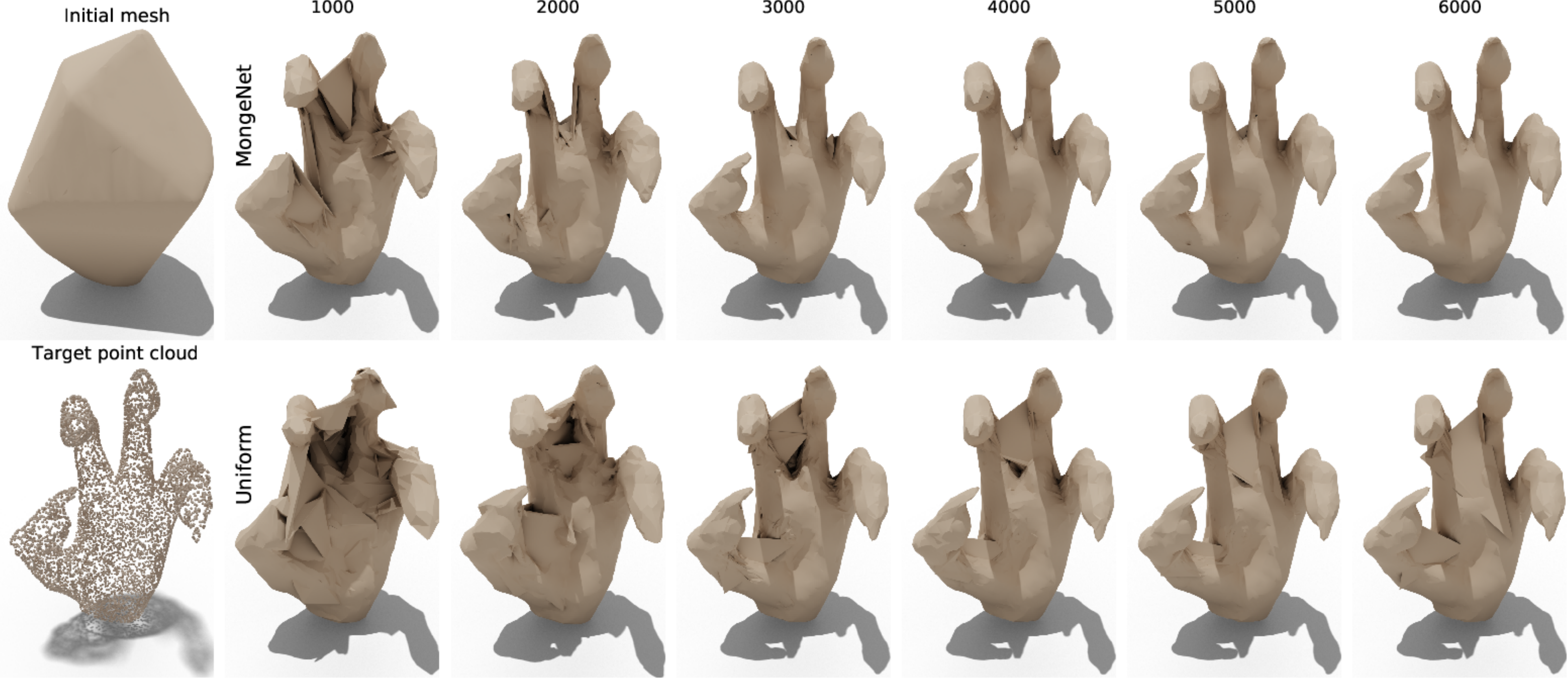}
\caption{\textbf{First row:} Loss function and metric evolution for 16 different random initializations. In blue the Point2Mesh method equipped with the MongeNet sampler and in red with the random uniform sampler, with translucent color the value for each individual run and in bold the average over all the runs.
\textbf{Second and third row:} Evolution of the optimized shape throughout the iterations for Point2Mesh equipped with MongeNet and with the random uniform sampler respectively.}
\label{fig:GalleryP2meshTraning}
\end{figure*}

We also conducted an in-depth analysis for an arduous mesh by running Point2Mesh 16 times, while monitoring all the metrics and the training loss, which are summarized in Figure~\ref{fig:GalleryP2meshTraning}. Across this experiment, all of the meta-parameters remain identical with the exception of the sampling technique. We noticed that for every scenario, because MongeNet provides a better representation for the face primitives, the distance between the target point cloud and the point cloud sample from the mesh decreased faster. As a result, for a given optimization time, the fidelity to the input point cloud was improved. At the end of the optimization, we can observe that only the MongeNet variant can recover the non-convex features of the hand.

\section{Conclusion}
In this paper, we have highlighted the limitations of the standard mesh sampling technique adopted by most of the 3D deep learning models including its susceptibility to irregular sampling and clamping, resulting in noisy distance estimation. To address this, we proposed a novel algorithm to sample point clouds from triangular meshes, formulated as an optimal transport problem between simplexes and discrete measures, for which the solution is swiftly learned by a neural network model named MongeNet. MongeNet is fast, fully differentiable, and can be adopted either for loss computation during training or for metric evaluation during testing. To demonstrate the efficacy of the proposed approach, we compared MongeNet to existing techniques widely used within the computer graphics community and evaluated the mesh approximation error using the challenging ShapeNet dataset. As a direct application, we also evaluated MongeNet on mesh approximation of noisy point clouds using the Point2Mesh backbone. In all these experiments, MongeNet outperforms existing techniques including the widely used random uniform sampling, for a modest extra computational cost.  

\section*{Acknowledgement}
This work was funded in part through an Australian Department of Industry, Energy and Resources CRC-P project between CSIRO, Maxwell Plus and I-Med Radiology Network.

{\small
\bibliographystyle{ieee_fullname}
\bibliography{egbib}
}

\end{document}